\documentclass[letterpaper,twoside,11pt]{article} 
\usepackage[style=apa,citestyle=authoryear,maxnames=10,maxcitenames=1,giveninits=true,backend=biber,uniquelist=false,uniquename=init,minbibnames=1]{biblatex}
\usepackage{jair, theapa, rawfonts}

\usepackage[utf8]{inputenc} 
\usepackage[T1]{fontenc}    
\usepackage{hyperref}       
\usepackage{url}            
\usepackage{booktabs}       
\usepackage{amsfonts}       
\usepackage{nicefrac}       
\usepackage{microtype}      
\usepackage{xcolor}         

\usepackage{xspace}
\usepackage{amsmath}
\usepackage{amsthm}
\usepackage{amssymb}
\usepackage{enumitem}
\usepackage[textsize=tiny,colorinlistoftodos,prependcaption]{todonotes}
\usepackage{booktabs}
\usepackage{subcaption}
\usepackage{comment}
\usepackage[capitalize]{cleveref}
\usepackage{float}
\newfloat{algorithm}{t}{lop}
\floatname{algorithm}{Algorithm}
\usepackage{tikz}
\usetikzlibrary{arrows.meta}

\providecommand{\ceil}[1]{\ensuremath{\left\lceil #1\right\rceil}}

\providecommand{\sas}{\ensuremath{\text{SAS}^{+}}\xspace}

\providecommand{\astar}{\ensuremath{\text{A}^{*}}\xspace}
\providecommand{\gbfs}{\ensuremath{\text{GBFS}}\xspace}

\providecommand{\hvalue}[1]{\ensuremath{h^{#1}}\xspace}
\providecommand{\hff}{\hvalue{\text{FF}}}
\providecommand{\hgc}{\hvalue{\text{GC}}}
\providecommand{\hstar}{\hvalue{*}}

\providecommand{\hnn}{$\hat h$\xspace}
\providecommand{\hnrsl}{$\hat h^{\text{N-RSL}}$\xspace}
\providecommand{\hboot}{$\hat h^{\text{Boot}}$\xspace}
\providecommand{\hgc}{\hvalue{\text{gc}}}

\providecommand{\hvfc}{\text{SUI}\xspace}
\providecommand{\hmin}{\text{SAI}\xspace}
\providecommand{\rw}{{RW}\xspace}
\providecommand{\bfs}{{BFS}\xspace}
\providecommand{\dfs}{{DFS}\xspace}
\providecommand{\bfsrw}{\text{FSM}\xspace}

\providecommand{\fssp}{{FS}\xspace}
\providecommand{\bssp}{{BS}\xspace}
\providecommand{\hnnrs}{$\hat h{^{20\%}_\text{\meanfx}}$\xspace}
\providecommand{\hnnrsfifty}{$\hat h{^{50\%}_\text{\meanfx}}$\xspace}

\providecommand{\hnnbase}{$\hat h_{0}$\xspace}
\providecommand{\hnnbfs}{$\hat h_{\text{bfs}}$\xspace}
\providecommand{\hnndfs}{$\hat h_{\text{dfs}}$\xspace}
\providecommand{\hnnrw}{$\hat h_{\text{rw}}$\xspace}
\providecommand{\hnnbfsrw}{$\hat h_\text{fsm}$\xspace}
\providecommand{\hnnbfsrwl}[1]{\ensuremath{\hat h_{#1}}\xspace}
\providecommand{\hnnnomutex}{\ensuremath{\hat h^{'}}\xspace}
\providecommand{\hnnnomutexl}[1]{\ensuremath{\hat h^{'}_{#1}}\xspace}

\providecommand{\define}[1]{#1}
\providecommand{\facts}{\ensuremath{F}\xspace}
\providecommand{\meanfx}{\ensuremath{\overline{F}}\xspace}
\providecommand{\default}{\ensuremath{200}\xspace}
\providecommand{\lforty}{\ensuremath{40}\xspace}
\providecommand{\distfarthest}{\ensuremath{d^*}\xspace}

\providecommand{\hnnrse}[1]{$\hat h{^{#1}_\text{\meanfx}}$\xspace}

\ifcsname dom\endcsname\else\DeclareMathOperator{\dom}{dom}\fi
\DeclareMathOperator{\pre}{pre}
\DeclareMathOperator{\eff}{eff}
\DeclareMathOperator{\sucs}{succ}
\DeclareMathOperator{\pred}{pred}

\DeclareMathOperator{\mutex}{mutex}

\ifcsname R\endcsname\else\newcommand{\R}{\ensuremath{\mathbb{R}}}\fi

\newtheorem{property}{Property}[section]

\colorlet{revcolor}{black}
\newcommand{\rev}[1]{\textcolor{revcolor}{#1}}

\jairheading{80}{2024}{243-271}{11/2023}{06/2024} 
\ShortHeadings{Understanding Sample Generation for Learning Heuristic Functions in Planning}
{Bettker, Minini, Pereira, \& Ritt}

\title{Understanding Sample Generation Strategies for\\
Learning Heuristic Functions in Classical Planning}

\author{%
  \name Rafael V.~Bettker\email rvbettker@inf.ufrgs.br    \\
  \name Pedro P.~Minini \email ppminini@inf.ufrgs.br      \\
  \name Andre G.~Pereira \email agpereira@inf.ufrgs.br    \\
  \name Marcus Ritt \email marcus.ritt@inf.ufrgs.br       \\
  \addr Universidade Federal do Rio Grande do Sul, Brazil \\
}

\begin{document}

\maketitle

\begin{abstract}
  We study the problem of learning good heuristic functions for classical planning tasks with neural networks based on samples represented by states with their cost-to-goal estimates.  The heuristic function is learned for a state space and goal condition with the number of samples limited to a fraction of the size of the state space, and must generalize well for all states of the state space with the same goal condition.  Our main goal is to better understand the influence of sample generation strategies on the performance of a greedy best-first heuristic search (GBFS) guided by a learned heuristic function.  In a set of controlled experiments, we \rev{find that two main factors determine the quality of the learned heuristic: the algorithm used to generate the sample set and how close the sample estimates to the perfect cost-to-goal are}. These two factors are dependent: having perfect cost-to-goal estimates is insufficient if the samples are not well distributed across the state space.  We also study other effects, such as adding samples with high-value estimates.  Based on our findings, we propose practical strategies to improve the quality of learned heuristics: three strategies that aim to generate more representative states and two strategies that improve the cost-to-goal estimates.  \rev{Our practical strategies result in a learned heuristic that, when guiding a GBFS algorithm, increases by more than 30\% the mean coverage compared to a baseline learned heuristic.}
\end{abstract}

\section{Introduction}
\label{sec:introduction}

A planning task is defined by the initial state that describes the starting conditions of the environment, the goal condition that specifies the conditions to be achieved, and the operators that can be applied to states of the planning task. Each operator has preconditions that must be satisfied to be applicable and effects that are the changes that occur in the state after the operator is applied. Planning is finding a sequence of operators that transforms an initial state into one that satisfies the goal condition. Applying an operator has a cost, and the cost of a plan is the total cost of its operators. A plan is optimal if its cost is minimal among all plans. One effective approach for solving planning tasks is to use algorithms from the best-first search family, which rely on a guiding function $f$ that estimates the quality of a state $s$. Best-first search expands first the state with the best quality. A heuristic function $h(s)$ estimates the cost-to-goal from state $s$ and may be used as a quality measure by $f$. One algorithm within the best-first search family is greedy best-first search (\gbfs)~\parencite{doran-michie-rsl1966}, which is only guided by the heuristic function $h(s)$. Another example is the \astar search algorithm, guided by $f(s) = g(s) + h(s)$~\parencite{hart-et-al-ieeessc1968}, where $g(s)$ is the current cost to reach the state~$s$ from the initial state. Generally, best-first search algorithms are more effective when the heuristic function~$h$ better estimates the optimal cost-to-goal, although this is not always the case~\parencite{Holte/2010}.

Some of the most successful heuristic functions~$h$ are based on methods that solve a simplified version of the planning task and use the cost to solve the simplified task to produce the cost-to-goal estimate. Examples of methods used to produce heuristic functions include delete relaxation~\parencite{bonet2001planning,Hoffmann.Nebel/2001}, landmarks~\parencite{hoffmann-et-al-jair2004,Helmert.Domshlak/2009,Karpas.Domshlak/2009}, critical paths~\parencite{haslum-geffner-aips2000}, constraints-based~\parencite{van2007lp,bonet-ijcai2013}, and abstractions~\parencite{Culberson.Schaeffer/1998,edelkamp2001,helmert2007flexible}. Many of these heuristics come with additional properties, such as admissibility. A heuristic is admissible if it does not overestimate the optimal cost-to-goal. \astar guided by admissible heuristic is guaranteed to find an optimal plan if one exists.

Increasing interest in learning heuristic functions with \define{neural networks}~(NNs) has been driven by rapid progress in other application areas. Some works in this area include those by~\textcite{samadi-et-al-aaai2008,Arfaee.etal/2011,Agostinelli.etal/2019,Ferber.etal/2020a,Shen.etal/2020,Yu.etal/2020,Toyer.etal/2020,Ferber.etal/2022,OToole/2022}. The basic approach is simple: generate a set of state pairs along with their estimated cost-to-goal values, then train a supervised model on these pairs. This model learns to predict cost-to-goal values for new, unseen states. \rev{Since learning heuristics that match the true cost-to-goal may not be optimal, an alternative method involves learning to rank states by optimizing the Kendall rank correlation coefficient during training, as in~\textcite{garrett2016learning}}. Nonetheless, a successful approach to planning has to solve several challenges:

\begin{enumerate}[label=C\arabic*),left=0pt]
\itemsep0pt
\item State spaces are implicitly defined and mostly exponential in the size of a compact description. Therefore, random samples are hard to generate and may be infeasible, unreachable from an initial state, or unable to reach the goal. Samples are usually generated by expanding the state space through forward (progression) or backward search~(regression).
\item Estimates of cost-to-goal are typically hard to obtain and finding the perfect cost-to-goal amounts to solving the task on the samples.
\item Planning domains are very different, and logic-based heuristics, which leverage the logical relationships encoded in the description of the planning task, apply to any domain. This results in the problem of transferring a learned heuristic to new domains, tasks, or state spaces.
\item Planners, which are systems designed to find solutions for planning tasks, rely on evaluating many states per second, so computing the heuristic function should be fast, or the learned heuristic must be more informed. However, there is a trade-off between a more informed learned heuristic and the complexity of the model.
\end{enumerate}

In this paper, we are interested in strategies for generating samples which we introduce in Section~\ref{sec:methods}, and in particular, what characterizes samples that lead to heuristic functions that perform well. To this end, Section~\ref{sec:experiments} presents a systematic study of the contributions of each strategy when solving distinct initial states of a single state space. We learn state-space-specific heuristics using a \define{feedforward neural network}~(FNN), focusing mainly on the quality of the learned heuristic and its influence on the number of expanded states and coverage. We aim to understand better how to learn high-quality heuristics. In experiments on small state spaces in Section~\ref{sec:experiment1}, we investigate the effect of different sampling strategies, the quality of the learned heuristic with an increasing number of samples, and the effect of a different subset of states part of the sample set on the learned heuristic; we also evaluate how the quality of the estimates of cost-to-goal influences the effectiveness of learned heuristic to guide a search algorithm. Then, in Section~\ref{sec:experiment2} we compare the best-proposed strategy with a baseline and traditional logic-based heuristics over large state spaces. Furthermore, we qualitatively compare existing methods in the literature in Section~\ref{sec:comparison-other-methods}. Finally, we summarize the main findings in Section~\ref{sec:conclusions}.

\subsection{Contributions}
\label{sec:contributions}
Through controlled experiments on planning tasks with small state spaces, we identify several techniques that improve the quality of the samples used for training. The contributions include:

\begin{itemize}
\item A sample generation algorithm that can better generate a representative subset of the state space through a combination of breadth-first search (expanding states close to the goal) followed by random walks from the breadth-first search's leaves (Section~\ref{sec:sampling-generation}).
\item State space-based estimations to limit the sampling regression depth to avoid large cost-to-goal overestimates (Section~\ref{sec:rollout-depth-limit}).
\item Two methods to improve cost-to-goal estimates based on detecting samples from the same or neighboring states (Section~\ref{sec:hvalue}).
\item A systematic study on sampling quality (Section~\ref{sec:experiments}).
\end{itemize}

\section{Preliminaries}
\label{sec:background}
This section introduces the foundational concepts and notations necessary for the discussions that follow. We also provide an overview of the existing literature related to our work.

\subsection{Background and Notation}

A~\sas planning task~\parencite{Backstrom.Nebel/1995} is a tuple $\Pi=\langle\mathcal{V},\mathcal{O},s_0,s^*, \text{cost}\rangle$, where $\mathcal{V}$ is a set of variables, $\mathcal{O}$ a set of operators, $s_0$~an initial state, $s^*$ the goal condition, and $\text{cost}:\mathcal{O}\rightarrow\R_{+}$ a function mapping operators to costs. A variable $v\in \mathcal{V}$ has a finite domain~$D(v)$, and a partial state~$s$ is a partial function $s:\mathcal{V}\rightarrow \mathcal{D}$, with $\mathcal{D}=\cup_{v\in \mathcal{V}}D(v)$ and such that $s(v)\in D(v)$, whenever $s(v)$ is defined. Otherwise, $s(v)$ is undefined, written as $s(v)=\bot$. The domain of $s$ is $\dom(s)=\{v\in \mathcal{V}\mid s(v)\neq\bot\}$. We also write $s|_U$ for the restriction of $s$ to a subset $U\subseteq\mathcal{V}$, i.e.~the state that agrees with $s$ on $U$ and is undefined otherwise. A (complete) state $s$ is a partial state with $\dom(s)=\mathcal{V}$. The initial state~$s_0$ is a state, and the goal condition~$s^*$ is a partial state. A partial state $s$ can \rev{also be interpreted as a set of facts $\{(v,s(v))\mid v\in\dom(s)\}$. We use these interpretations interchangeably, and their use will be clear from the context. We further write $\mathcal{S}(s)$ for the set of all complete states that agree with partial state $s$, namely $\mathcal{S}(s)=\{ s' \mid \dom(s')=\mathcal{V} \land s'|_{\dom(s)} = s\}$.} An operator $o \in O$ is a pair of preconditions and effects $(\pre(o), \eff(o))$, both partial states. \rev{Operator $o$ can be applied to state~$s$ if $s\supseteq \pre(o)$, and its application produces a successor state $s' = \sucs(s,o) := \eff(o) \circ s$, where $s'=t\circ s$ is defined by $s'(v) = t(v)$ for all $v\in\dom(t)$, and $s'(v) = s(v)$ otherwise.} The set of all successor states of state $s$ is~$\sucs(s)=\{\sucs(s,o)\mid o\in \mathcal{O}, \text{o applicable to }s\}$.  A sequence of operators $\pi=(o_1,\ldots,o_k)$ with $o_i\in \mathcal{O}$ is valid for initial state~$s_0$ if for $i\in[k]$ operator~$o_i$ can be applied to $s_{i-1}$ and produces $s_i=\sucs(s_{i-1},o_i)$. A \define{plan} is a valid sequence~$\pi$ for~$s_0$ such that~$s_k\supseteq s^*$ and the cost of plan~$\pi$ is $\sum_{i\in[k]} \text{cost}(o_i)$.

The predecessor $\pred(s,o)$ of a partial state $s$ under operator $o$ is obtained by \define{regression}. \rev{It is again a partial state, and all complete states in $\mathcal{S}(\pred(s,o))$ can reach $s$ by applying operator $o$.} For regression, we consider an operator $o\in \mathcal{O}$ to be relevant for partial state~$s$ if $\eff_r=\dom(\eff(o))\cap\dom(s)\neq\emptyset$; the operator is consistent if $s \subseteq \eff(o)|_{\eff_r}\circ\pre(o)$. Relevance requires that at least one variable is defined both in the effect and in the partial state to be regressed, and consistency requires an agreement on such variables or preconditions that are maintained. An operator~$o$ then is \define{backward} applicable in partial state~$s$ if it is relevant and consistent with~$s$ and leads to predecessor $\pred(s,o)=\pre(o)\circ (s|_{\dom(s)\setminus\eff_r})$. Note that $\sucs(\pred(s,o),o) \supseteq s$, but may differ from $s$. Similar to progression, a partial state~$s$ has predecessors $\pred(s)=\{\pred(s,o)\mid o\in \mathcal{O}, \text{o backward applicable to }s\}$. A regression sequence from state $s_0$ then is valid if $o_i$ can be applied to $s_{i-1}$ and produces $s_i=\pred(s_{i-1},o_i)$. All partial states~$s_i$ can reach a partial state $s\sup$. Since the goal is a partial state, a valid regression sequence $\rho=(o_1,\ldots,o_k)$ will generate a partial state that can reach the goal in at most $k$ steps and with a cost at most $\sum_{i\in[k]}\text{cost}(o_i)$.

Given a planning task, the set of all states over $\mathcal{V}$ is the state space. The \define{forward state space}~(\fssp) is the set of all states reachable from the initial state~$s_0$ by applying a sequence of operators. Similarly, the \define{backward state space}~(\bssp) is the set of all partial states reachable from the goal~$s^*$ by applying a sequence of backward applicable operators. Note that it is possible that the backward state space may contain partial states that cannot be reached by a forward search (also called spurious states, \rev{e.g.}~by \textcite{Alcazar.etal/2013}). This problem is addressed in \cref{sec:methods}.

\subsection{\rev{Representing and Generating Samples for Learning Heuristic Functions}}
\label{sec:learning}
A heuristic function~$h:\mathcal{S}\rightarrow\R_{+}\cup\{\infty\}$ maps each state in the state space~$\mathcal{S}$ to a non-negative \rev{cost-to-goal estimate~($h$-value), or to $\infty$ when the heuristic concludes that the state cannot reach the goal}. For example, the goal-count heuristic for a state~$s$ and goal condition $s^*$ computes the number of unsatisfied facts of the goal condition by the state i.e. $|\{ v \in \dom(s^*) \mid s(v)\neq s^*(v)\}|$ -- the resulting count indicates how far $s$ is from satisfying the goal condition.

Many heuristics for classical planning are derived from a model of the task, such as the \sas model introduced in the previous section. An obvious alternative is to learn to map a state $s$ to its heuristic value $h(s)$. We focus on learning with NNs, although other supervised learning methods could be used. To learn a heuristic function, an NN is trained on pairs of states~$s$ and cost-to-goal estimates~$c$. The learned heuristic functions are usually not admissible, so traditional optimality guarantees are lost.

A propositional representation of a state is more suitable for learning functions over states, as the variables in a planning task are categorical variables. To this end, consider a planning task $\Pi=\langle\mathcal{V},\mathcal{O},s_0,s^*, \text{cost}\rangle$, and let $\mathcal{V}=\{v_1,\ldots,v_n\}$ and $D(v_i)=\{d_{i1},\ldots,d_{i,z_i}\}$, $i\in[n]$ be some order of the variables and their domains. We represent a state $s$ by a sequence of facts $$\mathcal{F}(s)=(f_{11},f_{12},\ldots,f_{1,z_1},\ldots,f_{n1},f_{n2},\ldots,f_{n,z_n}),$$ where each fact $f_{ij}=[s(v_i)=d_{ij}]$ indicates if variable $v_i$ assumes value $d_{ij}$ in state $s$. Note that facts $\mathcal{F}_i=\{f_{i1},\ldots,f_{i,z_i}\}$ corresponding to variable $v_i$ satisfy the consistency condition $\sum_{f\in \mathcal{F}_i} f\leq 1$ since each variable assumes at most one value, and $\sum_{f\in \mathcal{F}_i} f=0$ only if $v_i$ is undefined. More generally, for any set of propositions $\mathcal{P}$ we write $\mutex(\mathcal{P})$ if $\sum_{p\in \mathcal{P}} [p]\leq 1$ must be satisfied in states of $\Pi$. Many planning systems can deduce mutexes from the description of the planning task $\Pi$~\parencite{Helmert/2009}; we will discuss and analyze their utility for sampling states later.
The target output for training may be the cost-to-goal estimates directly or some encoding of them.

An important aspect of sample generation is the degree of dependency on the domain model or the planning task.  We would like to learn in a \rev{black-box setting} where we interact with the planning task only by functions that allow accessing the initial state~$s_0$, the goal condition~$s^*$, and the applicable operators in a partial state. In this setting, we do not have access to the logical description of operators, but only to \rev{functions that return the successors or predecessors of a partial state~\parencite{Sturtevant2019}.}

For several reasons, the \rev{black-box} setting is interesting for studying the problem of learning heuristic functions. \rev{For example, it still can be applied if} the domain model is unavailable because a simulator or learned models generate successors and predecessors. Also, a \rev{black-box setting} can more easily be adapted than traditional logic-based heuristics to domain models with richer descriptions. Finally, \rev{we can address a question of \textcite{frances2017purely}, asking if a planner that has only access to the state structure and goals can approach the performance of planners that also have access to the logical description of the domain, in the specific case of only learning heuristics functions.}

Another aspect of sample generation is the cost of generating them.  This depends on the number of samples and the cost to generate each. In particular, we have the problem of deciding how many samples are required since, generally, only a very small part of the state space can be sampled.
\rev{Also, we have to decide which label to assign to each sample. The perfect heuristic $h^*$, that maps each state~$s$ to the cost of an optimal $s$-plan or $\infty$ if no such plan exists, would provide useful labels. However, even with samples labeled with perfect heuristic $h^*$, we expect the search performance to also depend on the region of the state space represented by the samples and how well the NN generalizes. In general, labeling samples with $h^*$ is impractical.}
Therefore, we are mainly interested in good heuristic estimates that can be generated fast. We analyze the influence of sample size and quality experimentally later. \rev{The generation of samples to produce heuristic functions is not restricted to learning heuristics with NN. For example, \textcite{haslum2007domain} generates samples by random walks from the initial state to evaluate which patterns should be used to produce a pattern database heuristic.}

Additionally, network architecture and sample generation depend on the range of tasks the learner intends to generalize. This may be a state space with a fixed goal condition, a planning domain, or an entire planning formalism. In the first case, the learned function has to generalize over a set of planning tasks defined by any pair of initial state $s_0$ part of a fixed \fssp and fixed goal $s^*$. In the second case, the learned function has to generalize over all tasks part of a domain. Finally, a learning-based heuristic that generalizes over a planning formalism is domain-independent.

\subsection{Related Work}
\label{sec:related}

\rev{There have been two main approaches to define the architecture of  neural network for learning heuristic functions. In the first, the architecture depends on the domain model (e.g.~uses information about preconditions and effects of operators), in the second the architecture is independent of the model.}

The usual setting \rev{in the first approach}~\parencite{Toyer.etal/2018,Toyer.etal/2020,Shen.etal/2020,Gehring.etal/2022,Stahlberg.etal/2022} is to train a NN with samples of small tasks of a domain and evaluate it on larger tasks of the same domain.  The networks trained can be general networks such as neural logic machines~\parencite{Dong.etal/2019} and graph neural networks~\parencite{Gori.etal/2005,Scarselli.etal/2008,Rivlin.etal/2020}, or networks proposed explicitly in the context of planning such as Hypergraph Neural Networks~\parencite{Shen.etal/2020} and Action Schema Networks~\parencite{Toyer.etal/2018}.  These networks require the logical description of the domain and the task to be instantiated and can \rev{typically be generalized across different state spaces of a domain, and sometimes across domains, as is the case for Hypergraph Neural Networks}. These approaches also help in understanding learning heuristics. For example, the main goal of \textcite{Stahlberg.etal/2022} is to understand the expressive power and limitations of learning heuristics.  The main limitation of these approaches is the strong dependence on the domain model and task description.

\rev{In the second approach}~\parencite{Ferber.etal/2020a, Yu.etal/2020, Ferber.etal/2022, OToole/2022} typically trains an FNN and evaluates the learned heuristic on a state space using tasks with the same goal and different initial states. These networks are trained with pairs of states and cost-to-goal estimates. \textcite{Ferber.etal/2020a} systematically study hyperparameters on the architecture of the FNN and found that their influence is secondary. They found that for a fixed architecture, two aspects significantly influence how informed the heuristic is: the subset of selected samples and the size of the sample set.

Furthermore, \textcite{Yu.etal/2020} and \textcite{OToole/2022} perform backward searches from the goal, the former with depth-first search and the latter with random walks. Both use the lowest depth in which the state was generated as cost-to-goal estimates. \textcite{Ferber.etal/2022} uses a combination of backward and forward searches~\parencite{Arfaee.etal/2011}. First, it generates new initial states with backward random walks and then solves them with a GBFS guided by a learned heuristic. The plans found provide the samples, and each sample is a state in the plan with the cost-to-goal estimate as its distance to the goal.  \textcite{OToole/2022} also proposed a method to generate samples that do not use expansions. This method includes randomly generated states in the sample set with a cost-to-goal estimate equal to the maximum value in the sample set plus one.  \textcite{OToole/2022} showed that this method substantially increases coverage.  The methods from the second set of approaches are highly independent of the domain model and planning task description and require low computational resources to generate samples and train the FNN. Also, despite having competitive results compared to logic-based heuristics, they are still unable to surpass the goal-count heuristic.

In particular, this work follows the second set of approaches since we aim to use minimal information from the task description. In~\cref{sec:comparison-other-methods}, we compare the proposed approach to \textcite{OToole/2022} and \textcite{Ferber.etal/2022} as they share the same NN architecture and dataset, with variations in the sampling and training procedures.

\section{Sample Generation}
\label{sec:methods}

We aim to investigate sample generation systematically.
Therefore, we focus on how two aspects of sample generation influence the performance of the learned heuristic to guide a search algorithm: the states $s_i$ in the state space included in the samples set and the quality of the estimates $h_i$ of samples with respect to the $\hstar$-value. Learning a heuristic function requires a set of samples $(s_1,h_1),\ldots,(s_N,h_N)$, where each sample $(s_i,h_i), i\in[N]$ consists of a state $s_i$ and a cost-to-goal estimate (or $h$-value) $h_i$.

We restrict this study to generalizing over planning tasks with initial states part of the same forward state space~(\fssp) and a fixed goal condition. We study \rev{black-box} approaches with access to predecessors and successors of partial states through a black-box function, to the goal condition, and to the domain of each variable. We also study approaches that have access to mutexes derived from a \sas model.
We address first, in Section~\ref{sec:generation}, the generation of states, and then, in Section~\ref{sec:hvalue}, the estimation of the cost-to-goal. In both sections, we discuss approaches from the literature and introduce new methods. The methods are a novel sampling strategy combining regression by breadth-first search with random walk, an adaptive regression limit, and two improvement methods for cost-to-goal estimates.

\subsection{Generation of States}
\label{sec:generation}

Unlike other domains in machine learning, where datasets of samples are often collected in real-world experiments and need to be manually annotated and curated, sample generation here is an algorithmic problem since we can generate the state space and compute cost-to-goal estimates. Approaches from the literature to generate the states include methods based on progression from one or more initial states, random sampling of the state space, or regression from the goal. In both progression and regression, one can apply different expansion strategies such as random walks, breadth- or depth-first searches, or combinations of more than one strategy such as bootstrapping~\parencite{Arfaee.etal/2011}. A problem in progression and random sampling is obtaining the cost-to-goal estimates. Without access to efficiently computable heuristic functions, or in \rev{black-box} approaches, these values have to be obtained by search, which can have a cost exponential in the size of the task. To remain less dependent on models than logic-based methods and also more general, we focus on regression for which an upper bound on the cost-to-goal is readily available, as discussed in Sections~\ref{sec:sampling-generation}~and~\ref{sec:rollout-depth-limit}. Since regression leads to partial states, the problem of generating complete states is addressed in Section~\ref{sec:sample-completion}. Random sampling is also discussed in Section~\ref{sec:random-samples-theory}.

\subsubsection{Sampling by Regression}
\label{sec:sampling-generation}

To generate samples, we expand \rev{partial} states through regression from the backward state space. Expansion methods include breadth-first search (\bfs), depth-first search (\dfs), random walks (\rw), as well as a combination of BFS with random walks. \rev{We discuss the completion of these partial states in Section~\ref{sec:sample-completion}.} A regression rollout is defined as a series of partial state expansions, and it stops if the last expanded partial state has no predecessors or is at the depth limit $L$. The sampling generation process stops if the number of required samples $N$ has been reached. Note that random walks can have multiple rollouts due to the depth limit $L$, while \bfs and \dfs only have one.

During expansion, we optionally use mutexes obtained from an analysis of the planning task -- in this case as computed by Fast Downward~\parencite{Helmert/2006} -- to discard partial states which cannot be completed to complete states without violating a mutex, as described in Section~\ref{sec:sample-completion}. We also discard repeated partial states for random walk rollouts, such that a single rollout never cycles, although the same partial state may be sampled several times in different rollouts. Starting from $h(s^*)=0$, a partial state $s'\in\pred(s,o)$ obtained by applying an operator $o$ backwards to partial state $s$ has a cost-to-goal estimate $h(s')=h(s)+\text{cost}(o)$. For samples \rev{$s$} that satisfy the goal condition \rev{($\mathcal{S}(s)\subseteq \mathcal{S}(s^*)$)} we reset the cost-to-goal estimate to $h(s)=0$. Partial states are added to the sample set when generated in the random walks and when expanded in BFS and DFS. In all methods, operators backward applicable to a partial state are applied in random order.

Different expansion strategies generate sample sets with varying frequencies of optimal distances from the goal. In our experience, good coverage of states close to the goal, such as those obtained by BFS or random walks, is useful, as is the greater depth obtained by DFS or random walks. However, random walks from the goal often sample states close to the goal multiple times, and DFS can lead to a concentration of distant samples from the goal.  Based on these observations, we propose a novel combination of BFS and random walks called \bfsrw that aims to have a good coverage close to the goal and a diverse set of samples from the remaining state space. \bfsrw has two phases. In the first phase, a fixed percentage $p_\bfsrw$ of the $N$ samples is generated by BFS. (The value of $p_{FSM}$ also serves as a constraint on computational resources, and the BFS process terminates when either the desired number of samples or the resource limit is reached first.) The \bfs expands a partial state from layer $k$ that generates $n$ partial states from layer $k+1$, and these partial states are sampled only if the current total samples plus $n$ are within $p_{\bfsrw}N$ samples; otherwise, no partial states are sampled and \bfs expands another partial state. Let $Q$ be the partial states of the set of samples that have not been expanded.  The second phase generates multiple random walk rollouts, each starting from a partial state in $Q$ chosen randomly with a complete replacement only after all partial states have been selected once. This is repeated until reaching $N$ samples in the sample set. During a random walk, partial states sampled in the BFS phase are not resampled.

\subsubsection{Maximum Regression Limit}
\label{sec:rollout-depth-limit}

A simple strategy to limit the sample generation depth is to use a fixed maximum limit $L$. \textcite{Yu.etal/2020} and \textcite{OToole/2022} used this strategy with respective limits of $L=200$ and $L=500$.  However, since tasks have state spaces with different maximum distances to the goal, a fixed limit is undesirable for generating a representative part of the state space. If the maximum regression limit $L$ overestimates the maximum distance-to-goal, the corresponding cost-to-goal estimates could be considerably larger than their respective \hstar-values because states could be generated through extremely non-optimal paths. If the maximum regression limit $L$ underestimates the maximum distance-to-goal, some part of the state space may never be generated.

The ideal regression limit for a given algorithm should allow it to generate a representative part of the state space without reaching states by extremely non-optimal paths. Let \distfarthest denote the distance from the goal to its farthest state part of the state space. For BFS, \distfarthest \rev{is an upper bound on the regression limit because at depth \distfarthest the whole state space has been explored}; for DFS and random walks, higher limits are required since they do not follow optimal paths. Since \distfarthest, in general, is unknown \rev{and hard to compute}, we propose two practical methods to define state-space-dependent maximum regression limits that aim to estimate \distfarthest.

The first method uses the number of facts $F = |\mathcal{F}(s_0)|$ to estimate \distfarthest. Since a set of facts represents each state, if one assumes that all operators have no precondition and change exactly one fact, one can reach any state in the state space from any other state by applying at most~$F$ operators.
However, operators can modify more than one fact at a time. We refine this approximation by considering the average number of facts changed by the task operators. The second method is defined as~$\bar F=\ceil{\facts/\overline{\eff}}$ where $\overline{\eff}=\sum_{o\in \mathcal{O}} |\eff(o)|/|\mathcal{O}|$, i.e.,~the number of facts per mean number of effects in the operators.

\subsubsection{Sample Completion}
\label{sec:sample-completion}

Regression sampling generates a set of partial states, while the NN is trained on and receives as input complete states during the search. Therefore, we evaluate three approaches to complete the undefined variables $\mathcal{U}=\mathcal{V}\setminus\dom(s)$ of a partial state $s$. The first approach is a random assignment. \rev{The method assigns to each undefined variable $v \in \mathcal{U}$ of a partial state} a random value in $\dom(v)$. This completion strategy \rev{can be applied in a black-box setting} since it uses \rev{information about the structure of a state}.

The second method is mutex-based and aims to avoid states that are impossible to reach during the search. For each partial state, the method processes the variables $\mathcal{U}$ in a random order and assigns to each of the variables in $\mathcal{U}$ a random value in $\dom(v)$ that does not violate the mutexes. We repeat this method for each partial state at most $10$\,K times. If after $10$\,K times this method does not produce a complete state that respects the mutexes, we still include the partial state in the sample set. In this case, we maintain all variables $v\in\mathcal{U}$ undefined, which sets all facts $\mathcal{F}_v$ for variables $v\in\mathcal{V}^u$ to false\footnote{Empirically this case is negligible since it occurs in about $0.1\,\%$ of the samples, in four of nine domains.}.

Since the set of mutexes we use is incomplete, the mutex-based method can still generate complete states that are impossible to reach during the forward search. Thus, we evaluate an ideal completion method to investigate the influence of generating only complete states that are reachable during the search. For each partial state~$s$, this method samples a random complete state from in the forward state space. Only for this method, if during the sample generation regression, a partial state~$s$ is generated such that no full state is in the forward state space, then~$s$ is considered an invalid predecessor. This ideal method can only be applied to small tasks where we can enumerate the complete \fssp of the initial state~$s_0$.

In all three methods, we do not check for repeated states after completion, i.e., two partial states can generate the same complete state. As a result, it is possible for two complete states to have different $h$-values, and we address this in~\cref{sec:hmin}.

\subsubsection{Randomly Generated Samples}
\label{sec:random-samples-theory}

\textcite{OToole/2022} have shown that adding randomly generated samples to a set of samples generated by expansion improves the performance of the search algorithm guided by the learned heuristic. They propose to set the cost-to-goal estimate for randomly generated samples to $L+1$ for a maximum regression limit of $L$. To study the effect of randomly generated samples, we include this method in this study. These samples are generated from fully undefined states using the mutex-based completion technique. If the generated state $s$ is already part of the sample set, i.e.,~$s = s_i$ for some $i\in[N]$ it receives cost-to-goal estimate $h_i$, otherwise the cost estimate $1+\max_{i\in[N]} h_i$ that is larger than all samples estimates.

\subsection{Improving Cost-to-Goal Estimates}
\label{sec:hvalue}

We start by observing that cost-to-goal estimates never underestimate the true cost-to-goal $h^{*}$, as follows.

\begin{property}
  \label{prop:hvalue}
  The cost-to-goal estimate $h(s)$ of a sample $s$ obtained by regression satisfies $h(s)\geq h^*(s)$.
\end{property}
\begin{proof}
  This follows because each estimate is witnessed by a plan. As observed in Section~\ref{sec:background}, a valid regression sequence $\rho=(o_1,\ldots,o_k)$ generates a sequence of partial states $s_i=\pred(s_{i-1},o_i)$, $i\in[k]$ starting from the goal $s_0$, such that $s_k$ can reach the goal in at most $k$ steps and with cost at most $\sum_{i\in[k]}\text{cost}(o_i)$. Furthermore, if $r=\pred(t,o)$ and $r'\in\rev{\mathcal{S}(r)}$ is a complete state, we have $\sucs(r',o)\in\rev{\mathcal{S}(t)}$. Therefore, for any complete state $s$ sampled from $s_k$, $o_k,o_{k-1},\ldots,o_1$ is a valid plan. Thus, $h(s)$ cannot be lower than the optimal cost $\hstar(s)$.
\end{proof}

In general, we expect better $h$-value estimates to lead to better-learned heuristics, and in turn to less expanded states during a search, but as previously noted this is not necessarily the case~\parencite{Holte/2010}. Therefore, we apply two procedures that improve the cost-to-goal estimates but maintain \cref{prop:hvalue}. The first, dubbed \hmin, minimizes estimates over repeated samples, and the second, \hvfc, over successors of samples.

\subsubsection{Improvement of Repeated Samples}
\label{sec:hmin}

In most state spaces, it is common for a state to be sampled in more than one random walk rollout, ending up with multiple duplicates with different estimates. Thus, for all sampled states $s$ we update each cost-to-goal estimate to the best estimate $h(s) = \min\{h_i \mid s=s_i, i\in[N]\}$ Since different partial states can generate identical complete states, the improvement is applied to \emph{exact} partial states as well as complete states. We call this procedure \emph{sample improvement} (\hmin). Choosing the minimum $h$-value is clearly sound since, in all cases, we have valid plans from a regression that witness these distances; for the same reason, \cref{prop:hvalue} still holds.

\subsubsection{Improvement over Successors}
\label{sec:hvfc}

Besides sampling the same states, it is common to sample states that are neighbors in the state space, particularly for states close to the goal. \rev{Information from neighboring states can be} used to improve the cost-to-goal estimates, as follows. 
\rev{First, for fast subset testing, we create an empty trie $T$, and for each partial state $s_i$, $i\in[N]$, we insert $s_i$ into $T$ keyed by its facts $(s(v))_{v\in\mathcal{V}}$.} 
\rev{Next, we build a graph $G=(V,A)$ with all sampled partial states $V=\{s_i\mid i\in[N]\}$, and $A=\emptyset$.}
For every pair of partial states $s,t\in V$ such that for some operator $o\in\mathcal{O}$ applicable to $s$ we have \rev{$\mathcal{S}(\sucs(s,o))\subseteq\mathcal{S}(t)$}, we add an arc $(s,t)$ of weight $w(s,t)=\text{cost}(o)$ to $A$. (Unlike in regression, if $\pre(o)$ mentions an undefined variable in $s$ then $o$ is not applicable.) 
\rev{We use the trie~$T$ to search for each successor in the partial states that are supersets. }
For partial states generated by regression, by construction, at least one such successor exists, except for the goal $s^*$.  Using graph $G$, we propagate the cost-to-goal estimate from each sampled state to its sampled predecessors. We iterate over each arc $(s,t) \in A$ and update the cost-to-goal estimate $h(s) = \min(h(s), h(t)+w(s,t))$. The process continues as long as there are updates. We call this procedure \emph{successor improvement} (\hvfc). As for \hmin, we only add valid transitions among partial states, so all distances are still witnessed by plans, and \cref{prop:hvalue} is maintained. 

\subsection{Training Set Generation Workflow}

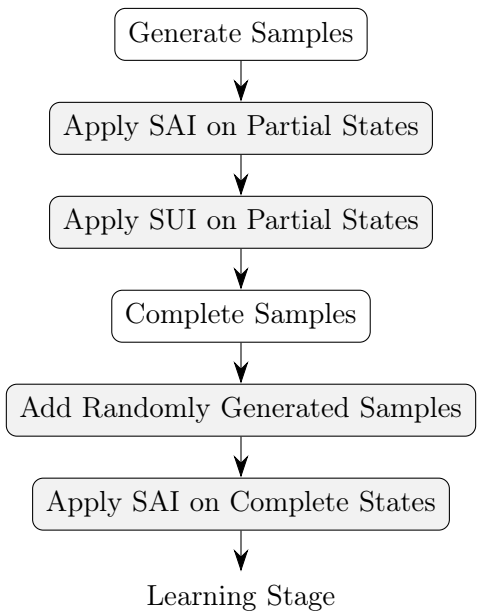
\begin{figure}[tb]
  \centering
  \begin{tikzpicture}
    \node[draw=black, fill=none, rounded corners, inner sep=5pt] (A) at (0,7.5) {Generate Samples };
    \node[draw=black, fill=black!5, rounded corners, inner sep=5pt] (B) at (0,6.25) {Apply SAI on Partial States};
    \node[draw=black, fill=black!5, rounded corners, inner sep=5pt] (C) at (0,5) {Apply SUI on Partial States};
    \node[draw=black, fill=none, rounded corners, inner sep=5pt] (D) at (0,3.75) {Complete Samples};
    \node[draw=black, fill=black!5, rounded corners, inner sep=5pt] (E) at (0,2.5) {Add Randomly Generated Samples};
    \node[draw=black, fill=black!5, rounded corners, inner sep=5pt] (F) at (0,1.25) {Apply SAI on Complete States};
    \node[fill=none, rounded corners, inner sep=5pt] (G) at (0,0) {Learning Stage};

    \draw[-{Stealth[length=3mm, width=2mm]}] (A) to (B);
    \draw[-{Stealth[length=3mm, width=2mm]}] (B) to (C);
    \draw[-{Stealth[length=3mm, width=2mm]}] (C) to (D);
    \draw[-{Stealth[length=3mm, width=2mm]}] (D) to (E);
    \draw[-{Stealth[length=3mm, width=2mm]}] (E) to (F);
    \draw[-{Stealth[length=3mm, width=2mm]}] (F) to (G);
  \end{tikzpicture}
  \caption{\rev{Training set generation workflow, shaded steps are optional.}}
  \label{fig:workflow}
\end{figure}

\rev{The process described in this paper follows the workflow illustrated in Figure~\ref{fig:workflow}.  It begins with the generation of samples using a sampling algorithm such as \bfsrw. The next steps improve the sample estimates using \hmin, followed by \hvfc, both of which are applied on partial states. The second and third steps are optional and can be applied independently. However, if applied, they are applied in this order. The fourth step uses a state completion technique to produce complete states from partial states. Then, optionally, we add randomly generated samples to the sample set. Finally, if \hmin was used previously, then it will be applied to the complete states. }

\label{sec:workflow}

\section{Experimental Setup}
\label{sec:experiments}

In this section, we present the experimental settings used throughout the experiments in small state spaces~(\cref{sec:experiment1}), where we can enumerate the complete forward state space with associated perfect cost-to-goal estimates \hstar, and in large state spaces~(\cref{sec:experiment2}), in order to validate our findings in a practical setting with large planning tasks.

\paragraph{Common Settings.}
\label{sec:methodology}

We use a \define{residual neural network} \parencite{He.etal/2016} to learn a heuristic for a state space. The network's input is a Boolean representation of the states, where a propositional fact is set to~$1$ if it is true in the state and~$0$ otherwise, as explained in Section~\ref{sec:learning}, and its output is a single neuron with the predicted $h$-value. The network has two hidden layers followed by a residual block with two hidden layers. Each hidden layer has~$250$ neurons that use ReLU activation and are initialized as proposed by \textcite{He.etal/2015}.
\rev{The output of the NN uses ReLU activation during training and evaluation.} The training uses the Adam optimizer~\parencite{Kingma.Ba/2015}, a learning rate of $10^{-4}$, an early-stop patience of~$100$, and a mean squared error loss function. Due to better results in preliminary experiments, we use batch sizes of $64$ for small and $512$ for large state spaces. We use $90\,\%$ of the sampled data as the training set, with the remaining $10\,\%$ as the validation set. \rev{The number of samples $N$ depends on the type of experiment and domain, and is given in the sections below.} In the experiments, different learned heuristics are denoted as \rev{$\hat h_B^A$, where $A$ and $B$ indicate different algorithmic choices}.

We select the domains and tasks from~\textcite{Ferber.etal/2022}, namely: Blocks World, Depot, Grid, N-Puzzle, Pipesworld-NoTankage, Rovers, Scanalyzer, Storage, Transport, and VisitAll. All domains have unit costs except for Scanalyzer and Transport, for which we consider the variant with unit costs. All methods are implemented on the Neural Fast Downward\footnote{Available at \url{https://github.com/PatrickFerber/NeuralFastDownward}.} planning system with PyTorch 1.9.0~\parencite{Ferber.etal/2020a,Paszke/2019}. The source code, planning tasks, and experiments are available\footnote{Available at \url{https://github.com/yaaig-ufrgs/NeuralFastDownward-FSM}.}. All experiments were run on a PC with an AMD Ryzen~9 3900X $12$-core processor, running at $4.2$~GHz with $32$~GB of main memory, using a single core per process distributed among $12$ (for small planning tasks) and $10$ cores (for large planning tasks). We solve all tasks with \gbfs guided by the learned heuristic~\hnn. When multiple states have the same heuristic value, the state that was generated earlier (i.e., has a lower generation order) is selected for expansion first (FIFO).

\paragraph{Initialization.}
\label{sec:nn_initialization}

An NN may fail to train if, after initialization, it outputs zero for all training samples~\parencite[this is called ``born dead'' in][]{Lu.etal/2020}. This condition arises when the weights and biases of the NN are initialized in a way that consistently maps the ReLU activation region to negative values, resulting in a zero gradient and no weight updates during training. Among \emph{all} small state space experiments using neural networks, about $5\,\%$ of the networks were born dead, where the Blocks World domain represented $84\,\%$ of the total of born deads, followed by VisitAll ($15\,\%$) and Transport (about $1\,\%$). In the large state space experiments, less than $1\,\%$ of the networks were born dead. Pipesworld-NoTankage represented $98\,\%$ of born deads, and Storage $2\,\%$. \rev{In the case of a born dead NN, we reinitialize it with a different seed until the NN outputs a non-zero value for some sample. In our experiments, no NN requires more than one reinitialization.}

\paragraph{Baseline.}
\label{sec:baseline}

We compare distinct configurations to a baseline \hnnbase similar to previous approaches from the literature. For the baseline, the NN is trained by a sampling method that uses random walks with a regression limit of~$200$ backward steps. In addition, mutexes are used during regression and for sample completion, but resetting the $h$-value to $0$ in goal states and the improvement strategies \hmin and \hvfc are turned off.

\paragraph{\bfs budget in \bfsrw.}
\label{sec:bfsrw_budget}

To determine a value for $p_{\bfsrw}$, preliminary experiments in small state spaces were performed using $p_{\bfsrw}\in\{0.01,0.05,0.1,0.2,\ldots,0.9\}$. We use the same baseline configuration but with the FSM sampling algorithm. The corresponding geometric mean expansions obtained were $64.35$, $57.89$, $56.31$, $60.43$, $57.6$, $57.3$, $62.39$, $69.22$, $77.49$, $78.09$, and $76.04$. respectively. Thus, we fix $p_{\bfsrw} = 10\,\%$ in the experiments.

\section{Experiments in Small State Spaces}
\label{sec:experiment1}

In this section, we study the behavior of different sampling methods on small state spaces. In particular, we analyze the effects of the following factors on the number of expanded states during search: the algorithms used to generate samples (partial states and their respective cost-to-goal estimates) (\cref{sec:experiment1-subset}), the partial state completion methods (\cref{sec:exp-state-completion}), and the accuracy of cost-to-goal estimates and methods to improve them (\cref{sec:exp-estimates-cost-to-goal}). Finally, we compare the proposed learning-based approaches to commonly used heuristics during search (\cref{sec:exp-comp-logic-based}).

For each domain, we select the task with the largest size state space that can be enumerated completely to obtain \hstar-values. We only select tasks with state spaces with $30$\,K states or more, and fewer than $1$\,M states.  Table~\ref{tab:small-instances} shows the tasks and their state space sizes. For domains Grid, Rovers, Scanalyzer, and Transport, the best task found had fewer than $30$\,K states, and VisitAll more than $1$\,M, so we manually modified these tasks. We could not find a non-trivial task within the limits for Depot, Pipesworld-NoTankage and Storage, so they were excluded from the experiments.  We generate the initial states for the small state spaces by performing a random walk of length~$200$ from the original initial state of a task. Rovers, Scanalyzer and VisitAll had duplicated initial states or states that satisfied the goal condition. Thus, we generate the initial states for these domains with random walks of length $25$, $50$, and $8$, respectively.

\begin{table}[tbp]
\centering
\caption{Size of the forward state spaces for the selected small tasks in seven domains. Tasks marked with~$*$ were modified.} 
\label{tab:small-instances}
\begin{tabular}{llrllr}
   \toprule
    Domain & Task & \#States & Domain & Task & \#States \\ 
    \midrule
    Blocks & blocks-7-0 & 65990 & Scanalyzer & p03$^*$ & 46080  \\
    Grid & prob01$^*$ & 452353 & Transport & p02$^*$ & 637632  \\
    N-Puzzle & prob-n3-1 & 181440 & VisitAll & p-1-4$^*$ & 79931  \\
    Rovers & p03$^*$ & 565824 & & & \\ 
   \bottomrule
\end{tabular}
\end{table}

In the small state space experiments, the coverage for all methods is $100\,\%$. Therefore, we use the number of expanded states to evaluate the quality of the heuristic function. In these experiments, we conducted tests using a total of $25$ combinations of sample seeds and network seeds ($5~\text{sample seeds} \times 5~\text{network seeds}$). Specifically, we trained five networks for each sample seed, ranging from sample seed $1$ to sample seed $5$, and repeated the process for each network seed. Additionally, we evaluated each network over $50$ different initial states for each domain. In other words, when we compare the number of expanded states, each cell in the tables represents the geometric mean of $25$ networks over $50$ test tasks, i.e., $1250$ searches with GBFS. The training time has been limited to $30$ minutes. If not stated otherwise, methods \bfs, \dfs, \rw and \bfsrw use mutexes, the improvement strategies \hmin and \hvfc, and the number of samples is equal to $1\%$ of the state space size. 
Under these conditions, more than $98.5\,\%$ of the NNs converge.

Furthermore, if not stated otherwise, for the fixed parameters we use the baseline setting defined in~\cref{sec:baseline} (regression limit $L=200$,  mutexes, no cost-to-goal improvement methods).

\subsection{Sample Generation}
\label{sec:experiment1-subset}

Our first set of experiments aims to analyze the influence of the sample generation methods on the quality of the learned heuristics.  \cref{sec:exp-sample-gen-algos} compares the different sampling algorithms, \cref{sec:exp-reg-limit} compares the adaptive regression limits, \cref{sec:exp-random-samples} investigates the effect of randomly generated samples.

In all the experiments below when comparing several treatments, we apply the non-parametric Mack-Skillings test for blocked designs with an equal number of replications in each experimental cell, with a confidence level $\alpha=0.01$ to see if the treatments are significantly different\footnote{The Mack-Skillings test is here preferable over, e.g., the more common Friedman test, since we have several replications per experimental cell.}, such . If not specified otherwise, the test is applied to each domain separately, with initial states as blocking factors. If treatments are significantly different we apply a corresponding post-hoc test to find the different treatments, again with a family-wise error rate of $\alpha=0.01$. For details about the tests see \textcite[ch. 7.9 and 7.10]{Hollander.etal/2014}.
\rev{In all following tables, the statistically best values are highlighted in bold.}

\subsubsection{Generation Algorithms}
\label{sec:exp-sample-gen-algos}
This experiment compares the four sample generation algorithms \bfs, \dfs, \rw, and \bfsrw and shows that a sample set including diverse regions of the forward state space (such as those generated by \rw and \bfsrw) yields fewer expansions in the geometric mean.

To control the effect of the cost-to-goal estimates on the quality of the learned heuristic, we replace sample estimates with optimal values $h^{*}$ before training. The left side of Table~\ref{tab:small-h-optimal} shows the geometric mean of the number of expanded states of a \gbfs guided by the learned heuristics. The right side in column FS shows the mean $h^{*}$-value of the whole forward state space, and columns BFS, DFS, RW, and FSM show the mean $h^{*}$-values of the sample sets representing $1\,\%$ of the forward state space~\fssp. We see that heuristic \hnnbfs leads to more expanded states than \hnndfs, which in turn expands about $40\,\%$ more states than \hnnrw and \hnnbfsrw, which perform similarly. All algorithms are significantly different for $\alpha=0.01$. Using \hnnbfs is significantly worse, and leads in all domains the highest or close to the highest number of expansions. Heuristic \hnndfs has a high number of expansions in Blocks World, N-Puzzle and Transport. Looking at the mean $h^{*}$-values, we see that samples generated by \bfs have the lowest, and those by \dfs the highest geometric mean estimates in all domains. Although the mean $h^{*}$-value of \dfs is closest to that of the whole forward state space \fssp, the resulting heuristic expands more states than \rw and \bfsrw, which generate states closer to the goal. Therefore, multiple random walk rollouts seem better than a single rollout with \bfs or \dfs.

\begin{table}[tb]
\centering
\caption{Comparison of sampling strategies \bfs, \dfs, \rw, and \bfsrw on $h^*$-values. Expanded states of \gbfs with learned heuristics, and mean \hstar-values over the entire forward state space (FS) and the generated sample sets.} 
\label{tab:small-h-optimal}
\begingroup\
\begin{tabular}{lrrrrrrrrr}
  \toprule  & \multicolumn{4}{c}{Expanded states} & \multicolumn{4}{c}{Mean $h^*$-values}\\ \cmidrule(lr){2-5} \cmidrule(lr){6-10}Domain & \hnnbfs & \hnndfs & \hnnrw & \hnnbfsrw & \fssp & \bfs & \dfs & \rw & \bfsrw \\ 
  \midrule
Blocks & 1881.33 & 63.24 & 34.46 & \textbf{30.76} & 18.77 & 10.77 & 17.68 & 11.93 & 14.42 \\ 
  Grid & 83.03 & 91.03 & \textbf{66.67} & \textbf{65.14} & 16.59 & 5.32 & 17.11 & 7.22 & 8.86 \\ 
  N-Puzzle & 929.73 & 132.37 & \textbf{85.02} & \textbf{83.95} & 21.97 & 10.42 & 20.17 & 20.02 & 19.78 \\ 
  Rovers & 11.63 & 12.20 & \textbf{11.08} & \textbf{11.09} & 6.45 & 2.34 & 5.19 & 4.95 & 4.99 \\ 
  Scanalyzer & 100.81 & 21.44 & \textbf{20.34} & \textbf{19.89} & 8.34 & 2.91 & 7.91 & 7.07 & 6.51 \\ 
  Transport & 94.69 & 39.96 & \textbf{18.81} & \textbf{18.42} & 12.23 & 2.94 & 11.30 & 10.05 & 9.53 \\ 
  VisitAll & 45.25 & 20.99 & \textbf{18.46} & \textbf{18.04} & 8.97 & 2.00 & 9.10 & 6.80 & 6.57 \\ 
   \midrule
Geo.~mean & 132.84 & 40.12 & 28.51 & \textbf{\rev{27.64}} & 12.21 & 4.22 & 11.45 & 8.81 & 9.09 \\ 
   \bottomrule
\end{tabular}
\endgroup
\end{table}

We now compare these results to results shown in Table~\ref{tab:small-h-estimate}, obtained on exactly the same states but using the cost-to-goal estimates obtained during sampling for training the NN. Note that the results for \bfs with estimated costs to the goal differ from those with exact values in Table~\ref{tab:small-h-optimal}. This happens because, during regression with \bfs, the cost-to-goal estimates are only exact on partial states; when turning them to complete states, the estimates can be larger than \hstar. Thus, \hnnbfs with the estimates obtained during regression is less informed.

We can see that the relative order of the methods concerning the number of expanded states remains the same, although all methods expand more states. Again, all algorithms have a different performance for $\alpha=0.01$. The increase in the number of expanded states is highest for \hnndfs, which expands \rev{a mean of $177.02$ states using $h$-values instead of $40.12$ when using $h^*$, i.e.~about} four times more states. In contrast, the other methods expand about twice as much, meaning that the estimates produced by \dfs during regression are inferior to those produced by the other methods.

Comparing the mean $\hstar$-values from Table~\ref{tab:small-h-optimal} to the mean $h$-values Table~\ref{tab:small-h-estimate}, we can see that the methods generate samples that overestimate $\hstar$-values and that \dfs overestimates much more than the others. Also, when contrasting the two extreme sample sets generated by \bfs and \dfs with \rw and \bfsrw, we find that in both Tables, \bfs generates samples closer to the goal and \dfs more distant from it.

In an additional experiment we verified that not only the absolute predicted $h$-values, but their ranking agrees well with $\hstar$-values. The mean correlation between these values, over all domains, are $0.302$, $0.237$, $0.464$, $0.521$, for methods BFS, DFS, RW, and FSM, respectively. Except for BFS, this agrees with the ranking of the methods according to the number of expanded states, and shows that a better sample distribution leads to predicted $h$-values that guide better.

Although \bfs has an estimation quality close to the \hstar-values, and a better correlation, it expands more states than the other methods. These results suggest that sampling more states in localized regions of the state space is not sufficient to achieve good results during search with \gbfs. Furthermore, \hnnbfsrw expands fewer states than \hnnrw and is the best in five of seven domains. Because \hnnbfsrw had a lower increase in expansions compared to \hnnrw, we focus on \bfsrw in the remaining experiments.

\begin{table}[tb]
\centering
\caption{Comparison of sampling strategies \bfs, \dfs, \rw, and \bfsrw on estimated $h$-values. Expanded states of \gbfs with learned heuristics, and mean \rev{\hstar-values over the entire forward state space (FS) and the mean estimated $h$-values over the generated sample sets.}} 
\label{tab:small-h-estimate}
\begingroup\
\begin{tabular}{lrrrrrrrrr}
  \toprule  & \multicolumn{4}{c}{Expanded states} & \multicolumn{4}{c}{Mean $h$-values}\\ \cmidrule(lr){2-5} \cmidrule(lr){6-10} 
Domain & \hnnbfs & \hnndfs & \hnnrw & \hnnbfsrw & \fssp & \bfs & \dfs & \rw & \bfsrw \\ 
  \midrule
Blocks & 1881.33 & 145.51 & 67.41 & \textbf{60.56} & 18.77 & 10.77 & 166.11 & 28.09 & 38.43 \\ 
  Grid & 271.15 & 1115.51 & \textbf{142.69} & \textbf{148.27} & 16.59 & 6.77 & 184.59 & 21.06 & 22.46 \\ 
  N-Puzzle & 929.73 & 707.33 & 173.90 & \textbf{146.11} & 21.97 & 10.42 & 187.51 & 96.29 & 90.65 \\ 
  Rovers & 49.45 & 22.27 & 18.63 & \textbf{16.62} & 6.45 & 4.65 & 27.11 & 25.90 & 24.91 \\ 
  Scanalyzer & 166.06 & 93.20 & 52.00 & \textbf{38.86} & 8.34 & 2.98 & 146.17 & 90.30 & 87.86 \\ 
  Transport & 166.52 & 567.07 & \textbf{89.84} & 103.13 & 12.23 & 3.27 & 196.12 & 95.20 & 88.59 \\ 
  VisitAll & 115.65 & 40.31 & \textbf{20.95} & \textbf{20.54} & 8.97 & 2.74 & 28.48 & 22.48 & 22.37 \\ 
   \midrule
Geo.~mean & 257.47 & 177.02 & 60.74 & \textbf{\rev{56.31}} & 12.21 & 5.14 & 103.50 & 43.30 & 44.38 \\ 
   \bottomrule
\end{tabular}
\endgroup
\end{table}

\subsubsection{Regression Limit}
\label{sec:exp-reg-limit}
In this experiment, we analyze the influence of the regression limit on the number of expanded states with sample generation strategy \bfsrw. The experiments show that values of regression limits \rev{slightly larger than} \distfarthest, which is the maximum distance to the goal in the state space, tend to yield fewer expansions in the geometric mean.

We compare the fixed regression limits $L=40$ (a good fixed limited for the small state spaces chosen by observing \distfarthest) and $L=200$ (used by \textcite{Yu.etal/2020}) with two state-space-dependent strategies that aim to estimate \distfarthest: setting the regression limit to the number of facts $F$ or to the number of facts divided by the mean number of effects $\bar F$. We also present results for $\nicefrac{\distfarthest}{2}$, $\distfarthest$ and $2\distfarthest$. The mean values for $F$, $\bar F$, and \distfarthest, are shown on the left-hand side in Table~\ref{tab:small-strategies-limits}. Both $F$ and $\bar F$ overestimate the mean largest distance \distfarthest (except for $\bar F$ in Blocks), which is desirable since random walks do not follow the optimal paths. Finally, $\bar F$ estimates \distfarthest better than $F$ for all domains.

\begin{table}[tb]
\centering
\caption{State space information and expanded states of \gbfs guided by \hnn trained on \bfsrw samples with different regression limits $L$ and no $h$-value improvements.} 
\label{tab:small-strategies-limits}
\begingroup\
\begin{tabular}{lrrrrrrrrrr}
   \toprule
   & \multicolumn{3}{c}{Mean Limits} & \multicolumn{7}{c}{Expanded states}\\ \cmidrule(lr){2-4} \cmidrule(lr){5-11}
Domain & \distfarthest & $F$ & $\bar F$ & $\nicefrac{\distfarthest}{2}$ & $\distfarthest$ & $2\distfarthest$ & \lforty &  \default & \facts & \meanfx \\ 
  \midrule
Blocks & 24 & 64 & 17 & 1279.70 & \textbf{43.48} & \textbf{45.29} & 49.37 & 60.56 & \textbf{46.81} & 89.89 \\ 
  Grid & 32 & 76 & 44 & \textbf{95.07} & 122.18 & 163.26 & 134.02 & 148.27 & 158.29 & 119.93 \\ 
  N-Puzzle & 31 & 81 & 41 & 112.78 & \textbf{66.37} & 77.42 & 75.29 & 146.11 & 84.25 & \textbf{66.52} \\ 
  Rovers & 19 & 32 & 27 & 32.05 & 17.93 & 15.73 & 15.75 & 16.62 & 15.69 & \textbf{14.76} \\ 
  Scanalyzer & 15 & 42 & 20 & 92.50 & 63.21 & 49.50 & \textbf{38.51} & \textbf{38.86} & \textbf{40.92} & 52.11 \\ 
  Transport & 17 & 66 & 35 & \textbf{48.91} & \textbf{47.95} & \textbf{51.16} & \textbf{49.63} & 103.13 & 65.24 & \textbf{49.39} \\ 
  VisitAll & 15 & 31 & 17 & 46.30 & 29.17 & 23.69 & \textbf{21.35} & \textbf{20.54} & 24.03 & 27.80 \\ 
   \midrule
Mean & 22 & 56 & 29 & 98.84 & 47.66 & 47.43 & \textbf{\rev{44.02}} & 56.31 & 48.47 & 49.78 \\ 
   \bottomrule
\end{tabular}
\endgroup
\end{table}

The right-hand side of Table~\ref{tab:small-strategies-limits} gives the number of expanded states for \gbfs guided by \hnn trained on \bfsrw samples with different regression limits $L$ and no $h$-value improvements. We see that limits \distfarthest, $2\distfarthest$, $L=40$, $F$, and $\bar F$ perform better than $\nicefrac{\distfarthest}{2}$ and the fixed limit $200$. The results with $\nicefrac{\distfarthest}{2}$ show that underestimating \distfarthest generally degrades performance. Also, note that $\bar F$ underestimates \distfarthest in Blocks, and the performance substantially degrades. The proposed strategies $F$ and $\bar F$ yield similar results to \distfarthest, $2\distfarthest$, and $L=40$. However, we do not have access to \distfarthest in general. Therefore, the proposed strategies provide good limits in practice and perform better than the previously used fixed limit of $L=200$.

\subsubsection{Random Samples}
\label{sec:exp-random-samples}
In this experiment, we evaluate the effect of adding randomly generated samples (as explained in Section~\ref{sec:random-samples-theory}) \rev{to a sample set generated using the sampling strategy \bfsrw and regression limit \meanfx. Unlike the previous experiments, we additionally apply the improvement strategies \hmin and \hvfc to the states sampled by regression (\hmin and \hvfc are discussed in Section~\ref{sec:exp-quality-estimates})}. The following experiments show that randomly generated samples are beneficial up to about $60\,\%$ of the total sample set, provided their cost-to-goal estimates are larger than the existing samples. We generate samples $S=\{(s_1,h_1),\ldots,(s_N,h_N)\}$ where $10\,\%, 20\,\%,\ldots,100\,\%$ are random samples (to which \hmin and \hvfc are not applied) and the rest is sampled with \bfsrw and a regression limit $\bar F$. Random samples get an $h$-value of $H+1$ where $H=\max_{i\in[N]} h_i$ is the largest $h$-value in samples $S$, except when they are part of the samples, in which case they receive the corresponding estimate (this happens in fewer than $1\,\%$ of the states). Note that when using $100\,\%$ of random samples, each has the cost-to-goal estimate equal to the regression limit $L+1$ instead of $H+1$, as we do not have samples in $S$.

\begin{table}[tb]
\centering
\caption{Expanded states of \gbfs with a learned heuristic over samples generated by \bfsrw with regression limit $\bar F$, all cost-to-goal improvement strategies, and a varying percentage of randomly generated samples.} 
\label{tab:small-contrasting}
\begingroup\
\begin{tabular}{lrrrrrrrr}
  \toprule  & \multicolumn{8}{c}{Percentage of random samples}\\ \cmidrule(lr){2-9}Domain & 0 & 10 & 20 & 30 & 40 & 50 & 60 & 70 \\ 
  \midrule
Blocks & 79.61 & \textbf{42.80} & \textbf{43.36} & 46.99 & 46.20 & \textbf{39.74} & 47.60 & 55.88 \\ 
  Grid & 70.14 & \textbf{50.98} & \textbf{49.37} & 59.90 & 53.27 & 56.63 & 59.28 & 86.81 \\ 
  N-Puzzle & \textbf{70.33} & \textbf{67.51} & \textbf{68.05} & \textbf{71.88} & \textbf{73.01} & \textbf{70.27} & 76.47 & 82.78 \\ 
  Rovers & 14.98 & 12.97 & \textbf{12.37} & \textbf{12.45} & \textbf{12.40} & 12.99 & 12.89 & 14.41 \\ 
  Scanalyzer & 37.00 & \textbf{24.47} & \textbf{22.40} & 24.57 & 26.13 & 30.00 & 24.04 & 26.61 \\ 
  Transport & \textbf{21.39} & \textbf{22.81} & 24.56 & 25.27 & 28.74 & 31.16 & 37.46 & 44.13 \\ 
  VisitAll & 28.54 & \textbf{21.57} & 20.76 & 21.05 & \textbf{21.02} & 21.96 & 22.75 & 25.04 \\ 
   \midrule
Geo.~mean & 38.82 & \rev{\textbf{30.22}} & \rev{\textbf{29.74}} & 31.81 & 32.11 & 32.95 & 34.40 & 40.21 \\ 
   \bottomrule
\end{tabular}
\endgroup
\end{table}

Table~\ref{tab:small-contrasting} shows the performance up to $70\,\%$ random samples. We have omitted $80\,\%$, $90\,\%$, and $100\,\%$ since expansions are higher~(respectively $45.12$, $56.85$, and $12081.97$). The number of expansions is considerably reduced when using random samples, with $20\,\%$ random samples performing slightly better than other percentages.

To better understand the effect of random samples, we have performed four additional experiments with $20\,\%$ of random samples. The first focuses on cost-to-goal estimates. We keep the samples but replace $H+1$ by small values, namely either a random $h$-value from the sample set $S$, or a random value drawn from $U[1,5]$. This leads to overall geometric means of $173.91$ and $3231.64$ expanded states, respectively. The second experiment forces the random samples to be part of the \fssp. This leads to a geometric mean of $31.11$ expanded states. The third experiment does not apply mutexes and leads to a geometric mean of $30.27$ expanded states. The fourth experiment generates random samples with true noise (each Boolean fact of a state has a $50\,\%$ chance of being $0$ or $1$), yielding a geometric mean of $45.01$ expanded states. From these additional experiments, it is clear that the most relevant factor is a high $h$-value, although generating random state samples with true noise yields more expansions than not using random samples. \rev{The most probable explanation for the effect of random samples seems to be that they enhance the likelihood of guiding the search toward samples where the network has learned good cost-to-go estimates, but that would require further experiments.}


\subsection{Partial State Completion}
\label{sec:exp-state-completion}
Now we focus on how sampled partial states are converted to complete states. In this experiment, \rev{we use \bfsrw, regression limit \meanfx,} and all the samples have optimal cost-to-goal estimates \hstar. We compare three different sample completion strategies for a partial state~$s$. All of them select a random state from the set of states \rev{$\mathcal{S}(s)$} represented by $s$ or a restriction of it: the set equals either to all states in $s$ (no restrictions), only those states that satisfy mutexes, or only states from the forward state space (perfect baseline). \rev{Note that even without restrictions, state completion is subject to the mutexes implicit in the \sas encoding.}

Table~\ref{tab:small-estimates-mutex} presents the expanded states \rev{and the probability of a completed state being in the \fssp} for these approaches. We can see that applying mutexes \rev{leads to a moderate reduction of the number of expanded states}, and is very close to an ideal completion of the states. However, completing randomly also presents competitive results, except for N-Puzzle and Blocks World. \rev{In precisely these two domains completing using mutexes is much better than randomly, since generating a valid combination of facts ``clear'' in Blocks World, or assigning the empty tile to the correct position is improbable.} \rev{For Rovers, Scanalyzer, Transport, and VisitAll, there is no distinction between the ``None'' and ``Mutex'' approaches because either there are no mutexes or the mutexes rarely apply to the sample states.}

\begin{table}[tb]
\centering
\caption{Expanded states of \gbfs with \hnn trained with \bfsrw, \meanfx, \hstar cost-to-goal estimates, and different state completion approaches\rev{, and the probability $P(s\in\fssp)$ of a completed state being in the forward state space.}}
\label{tab:small-estimates-mutex}
\begingroup\
\begin{tabular}{lrrrrr}
  \toprule  & \multicolumn{3}{c}{\rev{Expanded states}} & \multicolumn{2}{c}{\rev{$P(s\in\fssp)$}}\\
  \cmidrule(lr){2-4}\cmidrule(lr){5-6}
  Domain & None & Mutex & FS & None & Mutex \\ 
  \midrule
Blocks & 163.14 & 88.19 & \textbf{75.56} & 0.00 & 99.85\\
  Grid & 64.55 & 61.35 & \textbf{55.33} & 57.61 & 76.53\\
  N-Puzzle & 154.74 & 63.99 & \textbf{71.38} & 8.38 & 100.00\\
  Rovers & 10.77 & 10.98 & \textbf{10.10} & 24.26 & 23.27\\
  Scanalyzer & \textbf{30.34} & \textbf{31.10} & 33.64 & 100.00 & 100.00\\
  Transport & 17.23 & 17.20 & \textbf{16.12} & 7.30 & 7.50\\
  VisitAll & \textbf{20.93} & \textbf{21.18} & 21.89 & 64.41 & 63.83\\
   \midrule
Geo.~mean & 40.92 & 33.05 & \textbf{32.18} \\ 
   \bottomrule
\end{tabular}
\endgroup
\end{table}

\subsection{Estimates of Cost-to-Goal}
\label{sec:exp-estimates-cost-to-goal}
This set of experiments aims to analyze how techniques used to improve cost-to-goal estimates influence sample quality. To this end, in \cref{sec:exp-quality-estimates}, we directly compare the difference of the improved estimates (with \hmin and \hvfc) to \hstar-values. Then, to see how well the learned heuristics generalize, in \cref{sec:exp-eval-fs} we evaluate them over the complete forward state spaces from Table~\ref{tab:small-instances}.

\subsubsection{Quality of Estimates}
\label{sec:exp-quality-estimates}

First, we compare the quality of the cost-to-goal estimates to \hstar with and without $h$-value improvement techniques and distinct regression limits. \rev{We generate the samples with \bfsrw and varying regression depth limits, and complete the partial states with mutexes.} We demonstrate that using adaptive regression limits and cost-to-goal improvements (\hmin and \hvfc) consistently brings the estimates closer to \hstar. The results in Table~\ref{tab:small-samples-ssp} show the mean absolute difference between the sample estimates and \hstar, so smaller means indicate better approximations. Note that we are not evaluating the output of an NN, but the cost-to-goal estimates of the sample sets.

\begin{table}[tbp]
\centering
\caption{Mean difference of estimates of samples of the sample set to \hstar.}
\label{tab:small-samples-ssp}
\begin{tabular}{lrrrrrrr}
\toprule   & \multicolumn{3}{c}{No improvements} & \multicolumn{4}{c}{With \hmin and \hvfc}       \\
\cmidrule(lr){2-4}\cmidrule(lr){5-8}
Domain     & \default & \facts & \meanfx & \default & \facts & \meanfx \\
\midrule
Blocks     & 24.01    & 12.90  & 0.91    & 12.56    & 6.90   & \textbf{0.18}    \\
Grid       & 13.60    & 13.29  & 9.84    & 1.32     & 1.32   & \textbf{0.61}    \\
N-Puzzle   & 70.87    & 21.80  & 6.10    & 60.79    & 19.18  & \textbf{5.11}    \\
Rovers     & 19.92    & 11.58  & 9.74    & 6.70     & 5.29   & \textbf{4.88}    \\
Scanalyzer & 81.35    & 15.07  & 6.09    & 20.16    & 5.59   & \textbf{1.89}    \\
Transport  & 79.06    & 23.42  & 10.98   & 24.59    & 5.98   & \textbf{2.44}    \\
VisitAll   & 15.80    & 9.07   & 4.58    & 5.64     & 4.06   & \textbf{2.15}    \\
\midrule
Geo.~mean  & 33.45    & 14.56  & 5.56    & 10.95    & 5.35   & \textbf{\rev{1.60}}     \\
\bottomrule
\end{tabular}
\end{table}

The improvement strategies \hmin and \hvfc substantially reduce the estimates for all regression limiting methods. For regression limits $L$ of \default, \facts and \meanfx, using only \hmin reduces the estimates to $31.28$, $13.95$, $4.92$ respectively, and using only \hvfc reduces the estimates to $11.10$, $5.48$, $1.94$ respectively. Thus, \hvfc has the most effect on improving the cost-to-goal estimates compared to \hmin.
Also, the adaptive regression limiting methods are \rev{better than} the fixed default \default, and \meanfx has the best results. When comparing \default to \meanfx without $h$-value improvements, the cost-to-goal estimate difference to \hstar decreases by about $6$ times in the geometric mean, with Blocks World having the best performance, improving more than $25$ times. Finally, using both $h$-value improvements and a regression limit $\meanfx$ reduces the difference to \hstar from $33.45$ to only $1.60$.

\subsubsection{Evaluation Over The Forward State Spaces}
\label{sec:exp-eval-fs}

We now analyze the quality of the proposed learned heuristics and the logic-based heuristics fast-forward \hff~\parencite{Hoffmann.Nebel/2001} and goal-count \hgc over all states from the forward state space \fssp of each task. In summary, \hff better approximates the \hstar estimates over the whole forward state space, but the proposed approaches are close. Table~\ref{tab:small-strategies-sp-error} shows the results. Except for the baseline \hnnbase, the samples are generated with \bfsrw limited by an $L$ of \default, \facts or \meanfx, and improved with \hmin and \hvfc. The learned heuristic \hnnrs is the same as \hnnbfsrwl{\meanfx}, but $20\,\%$ of the samples are randomly generated. We see that \hnnbfsrwl{\meanfx} reduces the difference of the predicted result to the real one by about $11$ times when compared to \hnnbase, and when compared to \hgc it has the smallest difference in all but two domains. Also, \hnnbfsrwl{\meanfx} presents a similar mean difference to \hff. Note that due to the randomly generated samples in the sample set, \hnnrs doubles the difference compared to \hnnbfsrwl{\meanfx}. Blocks World is the only domain with a lower value, due to around two-thirds of the \fssp states having an \hstar-value within a range of two or less the value assigned for random samples, thus improving the average.

\begin{table}[tbp]
\centering
\caption{Mean difference of \hff, \hgc and \hnn, to \hstar when evaluated over the \fssp.}
\label{tab:small-strategies-sp-error}
\begin{tabular}{lrrrrrrr}
\toprule
Domain     & \hff          & \hgc          & \hnnbase & \hnnbfsrwl{\default} & \hnnbfsrwl{\facts} & \hnnbfsrwl{\meanfx} & \hnnrs        \\
\midrule
Blocks     & 6.76          & 13.37         & 26.46    & 16.58                & 9.84               & 2.91                & \textbf{2.42} \\
Grid       & 3.72          & 13.78         & 26.85    & 4.21                 & 4.10               & \textbf{2.73}       & 9.78          \\
N-Puzzle   & \textbf{4.19} & 14.86         & 79.84    & 65.37                & 23.90              & 6.75                & 12.73         \\
Rovers     & \textbf{0.17} & 3.31          & 11.08    & 3.18                 & 3.04               & 2.98                & 6.35          \\
Scanalyzer & 2.78          & \textbf{1.08} & 106.37   & 27.60                & 11.45              & 2.99                & 9.01          \\
Transport  & \textbf{1.13} & 8.63          & 109.77   & 33.53                & 12.54              & 7.05                & 14.89         \\
VisitAll   & \textbf{1.31} & 3.03          & 21.55    & 7.50                 & 5.56               & 2.21                & 4.74          \\ \midrule
Geo. mean    & \textbf{\rev{1.84}} & 5.92  &  39.80  & 13.91 & 8.13  & 3.57 & 7.40     \\
\bottomrule
\end{tabular}%
\end{table}

Comparing Tables~\ref{tab:small-samples-ssp} and \ref{tab:small-strategies-sp-error}, we observe that relative order between \default, \facts and \meanfx is preserved. The mean difference of the samples' estimates to \hstar for \meanfx is $1.60$ in Table~\ref{tab:small-samples-ssp}, and when the corresponding~\hnnbfsrwl{\meanfx} is required to generalize over the entire \fssp, the mean difference is $3.57$.

\subsection{Comparison to Logic-Based Heuristic Functions}
\label{sec:exp-comp-logic-based}

We now compare the NN-based heuristics with logic-based heuristics and show that the learned heuristics are competitive with commonly used heuristics. The number of expanded states of \gbfs guided by different heuristic functions is shown in Table~\ref{tab:small-samples-heuristic}. The NNs are trained with samples obtained with \bfsrw, all cost-to-goal improvement strategies and regression depth limited by \meanfx.

First, we see that \hnnbase expands fewer states than \hgc in most domains except Scanalyzer and VisitAll, but it is far worse than \hff except in Blocks World and VisitAll, where the learned heuristic has particularly good results. We also see that \hnnbfsrwl{\meanfx} expands fewer states than \hnnbase in five domains. In turn, \hff has better results than \hnnbfsrwl{\meanfx} in five domains. However, \hnnbfsrwl{\meanfx} surpasses \hff if $20\,\%$ of the samples are random, or if we increase the budget of \hnnbfsrwl{\meanfx} to $5\,\%$ (instead of $1\,\%$) of the number of states in the forward state space as shown in Table~\ref{tab:small-samples-pct}. This table also indicates that after increasing the budget to $50\,\%$ of the number of states in the forward state space, the gains in quality of the learned heuristic are negligible.

Comparing Tables~\ref{tab:small-strategies-sp-error} and~\ref{tab:small-samples-heuristic}, we see that for the NN-based heuristics, the order of approaches in terms of $h$--\hstar difference remains consistent for \hnnbase and \hnnbfsrwl{\meanfx}, but not for \hnnrs, which has a higher mean difference than \hnnbfsrwl{\meanfx} but presents the least state expansions, even when compared to \hff. With these results, we conclude that a better generalization over the forward state space is good for the samples obtained during regression. In contrast, despite worsening the mean difference to the \fssp, random samples are obtained after the regression procedure and can be helpful.

\begin{table}[tb]
\centering
\caption{Expanded states of \gbfs with different heuristic functions.} 
\label{tab:small-samples-heuristic}
\begingroup\
\begin{tabular}{lrrrrrr}
  \toprule Domain & \hstar & \hff & \hgc & \hnnbase & \hnnbfsrwl{\meanfx} & \hnnrs \\ 
  \midrule
Blocks & 19.25 & 136.35 & 248.06 & 67.41 & 79.61 & \textbf{43.36} \\ 
  Grid & 20.52 & \textbf{32.24} & 197.12 & 169.98 & 70.14 & 49.37 \\ 
  N-Puzzle & 22.40 & 121.17 & 607.62 & 173.82 & 70.33 & \textbf{68.05} \\ 
  Rovers & 9.80 & \textbf{10.54} & 44.61 & 18.63 & 14.98 & 12.37 \\ 
  Scanalyzer & 8.97 & \textbf{23.14} & 28.05 & 52.00 & 37.00 & \textbf{22.40} \\ 
  Transport & 13.17 & \textbf{17.12} & 184.37 & 89.57 & 21.39 & 24.56 \\ 
  VisitAll & 11.84 & 25.86 & \textbf{16.09} & 20.95 & 28.54 & 20.76 \\ 
   \midrule
Geo.~mean & 14.28 & 34.44 & 101.42 & 62.25 & 38.82 & \textbf{29.74} \\ 
   \bottomrule
\end{tabular}
\endgroup
\end{table}

\begin{table}[tb]
\centering
\caption{Expanded states of \gbfs with \hnnbfsrwl{\meanfx} trained with a number of samples corresponding to some percentage of the number of states in the \fssp of each task.} 
\label{tab:small-samples-pct}
\begingroup\
\begin{tabular}{lrrrr}
  \toprule Domain & 5\,\% & 25\,\% & 50\,\% & 100\,\% \\ 
  \midrule
Blocks & 25.40 & 20.35 & 20.03 & \textbf{19.68} \\ 
  Grid & 64.91 & 50.65 & 39.42 & \textbf{30.09} \\ 
  N-Puzzle & 34.72 & 26.44 & 25.07 & \textbf{23.73} \\ 
  Rovers & 14.78 & \textbf{13.36} & \textbf{13.26} & 17.43 \\ 
  Scanalyzer & 15.78 & \textbf{11.80} & \textbf{11.86} & 12.26 \\ 
  Transport & 19.19 & 17.49 & 16.20 & \textbf{15.32} \\ 
  VisitAll & 20.92 & \textbf{18.16} & 19.26 & 20.45 \\ 
   \midrule
Geo.~mean & 24.54 & 20.18 & \textbf{19.23} & \textbf{19.14} \\ 
   \bottomrule
\end{tabular}
\endgroup
\end{table}

\section{Experiments in Large State Spaces}
\label{sec:experiment2}

The main goal of the following experiments is to verify the findings from the previous sections on large state spaces, so we compare different configurations of the improved methods with logic-based heuristics and a baseline. In summary, the following experiments confirm that the previous findings also apply to large state spaces. Then, in \cref{sec:comparison-other-methods}, we compare the proposed approach with other learning-based approaches from the literature. We report results over $9$ combinations of sample seeds and network seeds ($3$ sample seeds and $3$ network seeds).

We use the benchmark defined by \textcite{Ferber.etal/2020a,Ferber.etal/2022}. \textcite{Ferber.etal/2020a} selects for each domain, IPC planning task with their original initial states, that are solved within $1$ and $900$ seconds by GBFS with \hff.  Each domain has the following number of selected tasks: Blocks World, $5$; Depot, $6$; Grid, $2$; N-Puzzle, $8$; Pipesworld-NoTankage, $10$; Rovers, $8$; Scanalyzer, $6$; Storage, $4$; Transport, $8$; VisitAll, $6$.  For each selected task, \textcite{Ferber.etal/2020a} generates new initial states with random walks from the original initial state. These new initial states with the original goal conditions of the selected tasks define the test planning tasks. For selected task \textcite{Ferber.etal/2022} experiments with $50$ moderate tasks resulting in a benchmark with $3,150$ planning tasks. We use the same benchmark used by \textcite{Ferber.etal/2022}.

We generate samples within $1$~hour and set $1$~hour as the maximum training time. Each of the $50$~initial states must be solved separately with \gbfs within $5$~minutes and $2$\,GB~RAM. We fix the number of samples at $N = 16\text{M} / |\mathcal{V}|$, \rev{such that domains with more variables $\mathcal{V}$ receive proportionally less samples. This} results in a mean of $500$\,MB RAM during sampling and $2$\,GB during the $h$-value improvement.

First, we reassess the previous results using the regression limits \facts and \meanfx on large state spaces since the previous experiments (Section \ref{sec:experiment1-subset}) produced similar results. Table~\ref{tab:large-instances-moderate-nomutex} shows the mean coverage and number of expanded states for the methods using \facts or \meanfx, with all $h$-value improvements, and with and without mutexes (denoted by \hnnnomutex).

\begin{table}[tbp]
\centering
\caption{Mean coverages and expanded states of the learned heuristics with both regression limit and their respective approaches not using mutexes (\hnnnomutex). Expanded states consider only the initial states solved by all heuristics; Grid, N-Puzzle and Storage had no common solved initial state. Geometric mean is used for the overall mean of expanded states.}
\label{tab:large-instances-moderate-nomutex}
\begin{tabular}{lrrrrrrrr}
    \toprule & \multicolumn{4}{c}{Coverage (\%)} & \multicolumn{4}{c}{Expanded states} \\
    \cmidrule(lr){2-5} \cmidrule(lr){6-9}
    Domain & \hnnbfsrwl{\facts} & \hnnbfsrwl{\meanfx} & \hnnnomutexl{\facts} & \hnnnomutexl{\meanfx} & \hnnbfsrwl{\facts} & \hnnbfsrwl{\meanfx} & \hnnnomutexl{\facts} & \hnnnomutexl{\meanfx} \\
    \midrule
    Blocks & 93.33 & 96.31 & \textbf{100.00} & \textbf{100.00} & 19452 & \textbf{2722} & 4241 & 5406 \\
    Depot & 81.26 & 84.67 & 85.96 & \textbf{86.41}             & \textbf{11869} & 13440 & 15528 & 14557 \\
    Grid & 76.56 & \textbf{81.11} & 41.33 & 38.33              & - & - & - & - \\
    N-Puzzle & 21.64 & \textbf{95.50} & 7.42 & 35.33           & - & - & - & - \\
    Pipes-NT & 17.80 & 17.78 & 17.82 & \textbf{18.07}          & \textbf{3157} & 7628 & 5414 & 5345 \\
    Rovers & \textbf{13.92} & 13.33 & 13.67 & 13.53            & \textbf{15} & 17 & 16 & \textbf{15} \\
    Scanalyzer & 66.44 & \textbf{67.78} & 66.15 & 66.67        & 146 & \textbf{35} & 188 & 38 \\
    Storage & 3.94 & \textbf{9.11} & 1.67 & 8.33               & - & - & - & - \\
    Transport & 75.53 & 85.97 & 81.42 & \textbf{87.50}         & 64985 & 32603 & 36064 & \textbf{29671} \\
    VisitAll & \textbf{92.81} & 80.52 & 92.70 & 79.00          & 1931 & 2978 & \textbf{1324} & 4575 \\
    \midrule
    Mean & 54.32 & \rev{\textbf{63.21}} & 50.81 & 53.32       & 2710 & \rev{\textbf{1849}} & 2152 & 2020 \\
    \bottomrule
\end{tabular}
\end{table}

When comparing the learned heuristic \hnnbfsrwl{\meanfx} over \hnnbfsrwl{\facts}, we see a mean coverage improvement of about $9\,\%$. All domains are improved or have very similar results, except VisitAll, where limiting the regression limit by \facts is better -- this is also observed in the small state space experiments. Without mutexes, the coverage improvements from \hnnnomutexl{\meanfx} over \hnnnomutexl{\facts} are minor. The smaller number of expanded states in approaches with \meanfx indicate samples of higher quality. With or without mutexes, using \meanfx has the highest positive effect in N-Puzzle, increasing its coverage by about four times. Also, we see that not using mutexes improves the results in Blocks World, Depot and Transport, while having a minimal effect in Pipesworld-NoTankage, Rovers, Scanalyzer and VisitAll. From these results, we conclude that \meanfx has better performance than \facts for large state spaces. Therefore, the following experiments will use \meanfx.

Next, we compare the logic-based heuristics \hff and \hgc, the baseline \hnnbase and the best approach \hnnrs. The results are presented in Table~\ref{tab:large-instances-moderate-cmp}.
We see that \hff dominates in most domains, achieving twice the mean coverage of the baseline \hnnbase. However, \hnnrs has only $12\,\%$ less mean coverage than \hff, improving \hnnbase by about $31\,\%$, with competitive coverage in most domains. Note that \hnnrs achieves better mean coverage than \hgc, with higher or equal coverage in $6$ out of $10$ domains. Also, in all domains except Pipesworld-NoTankage and Transport, the best-learned heuristic expands fewer states on the same initial states when compared to \hff, indicating that the learned heuristic is more informed and that the inferior coverage is an effect of the slower expansion speed of the NN-based heuristics. Note, however, \rev{since only tasks solved by all methods are considered, the evaluation is biased towards easier tasks}. Furthermore, when limiting \hff by the same number of expansions as the learned heuristic, \hff achieves coverage of $81.20$, meaning that it still excels in most states. Because the dataset used was generated from tasks that are solvable by \hff within $900$ seconds, the results are also biased towards better performance with a search guided by \hff.

\begin{table}[tbp]
\centering
\caption{Mean coverages and expanded states of the logic-based heuristics \hff and \hgc compared to the baseline learned heuristic \hnnbase and the best learned heuristic \hnnrs, obtained via training over samples with \bfsrw, \meanfx, $20$\,\% of random samples, and all $h$-value improvement strategies. Expanded states consider only the initial states solved by all heuristics; N-Puzzle and Storage had no common solved initial state. \rev{For expanded states we use the geometric mean.}}
\label{tab:large-instances-moderate-cmp}
\begin{tabular}{lrrrrrrrr}
    \toprule & \multicolumn{4}{c}{Coverage (\%)} & \multicolumn{4}{c}{Expanded states} \\
    \cmidrule(lr){2-5} \cmidrule(lr){6-9}
    Domain & \hff & \hgc & \hnnbase & \hnnrs & \hff & \hgc & \hnnbase & \hnnrs \\
    \midrule
    Blocks & \textbf{100.00} & \textbf{100.00} & \textbf{100.00} & \textbf{100.00} & 12499 & 32115 & 2771 & \textbf{2099} \\
    Depot & \textbf{94.33} & 80.00 & 57.19 & 89.26 & 44811 & 197999 & 13139 & \textbf{6756} \\
    Grid & \textbf{94.00} & 51.00 & 38.11 & 60.33 & 4579 & 23901 & \textbf{2313} & 4332 \\
    N-Puzzle & \textbf{92.50} & 4.00 & 13.75 & 86.81 & - & - & - & - \\
    Pipes-NT & 63.40 & 89.40 & 13.51 & \textbf{79.84} & \textbf{100} & 1348 & 7653 & 756 \\
    Rovers & \textbf{85.50} & 66.00 & 13.53 & 15.39 & 12 & 38 & 36 & \textbf{11} \\
    Scanalyzer & \textbf{100.00} & \textbf{100.00} & 59.70 & 73.67 & 50 & 81 & 1085 & \textbf{18} \\
    Storage & \textbf{33.00} & 13.50 & 1.94 & 27.67 & - & - & - & - \\
    Transport & \textbf{100.00} & \textbf{100.00} & 48.89 & \textbf{100.00} & \textbf{3505} & 244299 & 76457 & 7476 \\
    VisitAll & 92.00 & \textbf{100.00} & 74.19 & 98.85 & 31560 & \textbf{259} & 3787 & 1656 \\
    \midrule
    Mean & \rev{\textbf{85.47}} & 70.39 & 42.08 & 73.18 & 1810 & 5140 & 3582 & \rev{\textbf{881}} \\
    \bottomrule
\end{tabular}
\end{table}

When comparing Tables~\ref{tab:large-instances-moderate-nomutex} and \ref{tab:large-instances-moderate-cmp}, we notice that all NN-based heuristics have similarly poor results in Rovers, independent of configuration. Considering only the learned heuristics, when using $20\,\%$ of random samples (\hnnrs) instead of $0\,\%$ (\hnnbfsrwl{\meanfx}), there are intermediate improvements of about $15\,\%$ in Storage, Transport and VisitAll, and a significant improvement in Pipesworld-NoTankage, from approximately $18\,\%$ to $80\,\%$ coverage.

\subsection{Comparison to Other Sample-Based Methods}
\label{sec:comparison-other-methods}

\rev{In this section we compare the learned heuristics to those of \textcite{Ferber.etal/2022} and \textcite{OToole/2022}. All methods share the same benchmark and NN architecture, use a Boolean representation for the samples, and mutexes to complete unassigned state variables. They differ, however, in batch size, patience value, NN initialization functions, and percentages of data split into training and validation sets. Furthermore, machine configurations, libraries, and resource limits are different, and existing results cannot be easily reproduced. Thus, we limit ourselves to a qualitative comparison. Even within these constraints, the following experiments indicate that the proposed approach achieves better coverage in less time.}

\textcite{Ferber.etal/2022} learns using bootstrapping~\parencite{Arfaee.etal/2011}, which iteratively improves the learned heuristic by training with samples with cost-to-goal estimates of increasing distances from the goal.  The Bootstrap method generates new initial states with backward random walks from the goal for a given state space. For each generated initial state, the method attempts to find a plan with \gbfs guided by the current learned heuristic. The plans found with their respective states and cost-to-goal estimates are used as training samples. If the method finds a plan for more than $95$\% of the initial states, it generates initial states with longer random walks. \textcite{Ferber.etal/2022} perform sampling and training for up to $28$ hours, with a search time-limit of $10$ hours.

\textcite{OToole/2022} generates samples using random walks from the goal. They perform $5$ rollouts with a regression limit of $L=500$ and use the current depth as cost-to-goal estimates; they use the Tarski framework~\parencite{tarski:github:18} to perform the regression procedure and \rev{develop} a cost-to-goal improvement strategy equivalent to \hmin, and $50\,\%$ of the samples are randomly generated as described in Section~\ref{sec:random-samples-theory}. \textcite{OToole/2022} spend an unreported amount of time to generate $100$\,K samples and an average of $23$ minutes in training, with a search time-limit of $6$ minutes.

We now compare the coverage results between all methods over the same tasks. \rev{In the comparison we consider only the best configurations of both \textcite{Ferber.etal/2022} and \textcite{OToole/2022}. We also use $100$\,K samples to make them more comparable to \textcite{OToole/2022}.} In these experiments, our sampling and training procedures can take a combined time of up to $2$ hours, and we use a search time limit of $5$ minutes.

Table~\ref{tab:large-instances-moderate-100k} shows the coverage results of our best method using $0\,\%$, $20\,\%$ and $50\,\%$ of random samples, \rev{together with the results as reported by \textcite{Ferber.etal/2022} (\hboot) and \textcite{OToole/2022} (\hnrsl)}. Considering coverages, we notice that the heuristics \rev{\hnnrse{\cdot\%}} have results more similar to \hnrsl than to \hboot, and higher coverage, except for Grid, Rovers, Scanalyzer, and Storage. \rev{Heuristic \hboot is best} in Grid, Rovers, and Storage. Particularly in Storage, \hboot has $89\,\%$ coverage, \rev{while coverage of} the other approaches \rev{is} close to $20\,\%$. \rev{Heuristics \hnnrse{\cdot\%}} have the best coverage in six out of ten domains, namely Blocks World, Depot, N-Puzzle, Pipesworld-NoTankage, Transport, and VisitAll. All approaches have comparatively low coverage in Rovers, and \hboot and \hnrsl have low coverage in N-Puzzle when compared to the proposed approaches.

Generating samples only through regression (i.e., without solving states) and training \rev{afterward} is faster \rev{than} bootstrapping, as the states generated by the backward random walk must be solved with the currently learned heuristic to produce plans used as samples. Both \hnrsl and the proposed methods suggest that sampling using regression with improvement strategies (such as \hmin, \hvfc, and random samples) gives competitive results on most domains.

\begin{table}[th]
\centering
\caption{Mean coverage results of \textcite{Ferber.etal/2022} (\hboot) and \textcite{OToole/2022} (\hnrsl), with results obtained from their respective papers, and our best learned heuristic trained with $100$\,K samples, from which $0$\,\%, $20$\,\% and $50$\,\% were randomly generated.}
\label{tab:large-instances-moderate-100k}
\begin{tabular}{lrrrrrr}
    \toprule
    Domain & \hboot & \hnrsl & \hnnbfsrwl{\meanfx} & \hnnrs & \hnnrsfifty \\
    \midrule
    Blocks      & 0.00 & 91.50 & 99.96  & \textbf{100.00} & \textbf{100.00} \\
    Depot       & 32.00 & 58.80 & 80.56  & \textbf{91.52}  & 81.78  \\
    Grid        & \textbf{100.00} & 60.30 & 59.89  & 57.56  & 54.44  \\
    N-Puzzle    & 27.00 & 18.90 & 70.94  & 85.92  & \textbf{86.36}  \\
    Pipes-NT    & 36.00 & 69.60 & 19.00  & 91.51  & \textbf{96.62}  \\
    Rovers      & \textbf{36.00} & 12.50 & 13.47  & 14.33  & 14.50  \\
    Scanalyzer  & 33.00 & \textbf{94.10} & 67.00  & 66.67  & 66.15  \\
    Storage     & \textbf{89.00} & 16.40 & 19.67  & 22.00  & 19.17  \\
    Transport   & 84.00 & 70.80 & \textbf{100.00} & 99.67  & 86.89  \\
    VisitAll    & 17.00 & 95.40 & 84.52  & 99.04  & \textbf{99.70}  \\
    \midrule
    Mean & 45.40 & 58.80 & 61.50 & \rev{\textbf{72.82}} & 70.56 \\
    \bottomrule
\end{tabular}
\end{table}

According to \textcite{OToole/2022}, the proportion of random samples in the final sample set has the most positive effect on coverage -- approximately doubling it when going from $0\,\%$ to $50\,\%$ of random samples ($34.7$ vs.~$59.9$, from their supplementary material). As seen in Table~\ref{tab:large-instances-moderate-100k}, we also notice an improvement from using random samples, although smaller. The experiments show that all domains either improve or have similar results, and the mean coverage improves by about $10\,\%$. Also, except for Pipesworld-NoTankage, we saw no improvements above $5\,\%$ using $50\,\%$ of random samples compared to $20\,\%$. This means that despite having the highest coverage with $20\,\%$, using $50\,\%$ of random samples can maintain similar coverage while improving sampling time, as random samples are generated much faster than in regression.

\section{Conclusion}
\label{sec:conclusions}

We have presented a study of sample generation and improvement strategies for training NNs to learn heuristic functions for classical planning. We have revised existing approaches to sample generation and proposed a new strategy that uses regression with breadth-first search and random walks, as well as several techniques that improve cost-to-goal estimates. A systematic analysis using complete information of small state spaces indicates that: for the samples obtained through regression with states covering diverse portions of the state space without repeated samples close to the goal works best, and enough samples of good quality translate to good search performance that can be compared to logic-based heuristics. Also, we confirm the results from \textcite{OToole/2022} showing that having randomly generated samples up to a limit in the final sample set is positive.  Among the contributions, the $h$-value improvement strategy \hvfc and the adaptive regression limit \meanfx have the most positive effects on sampling quality. The former improves the accuracy of cost-to-goal estimates by analyzing the successors of a state, while the latter avoids overestimates by limiting the maximum regression limit.

\section*{Acknowledgments}

R.~V.~B. and P.~P.~M have contributed equally to this work.  We thank the Fast Downward planning system team, and especially Patrick Ferber, Malte Helmert, and J{\"o}rg Hoffmann for sharing the source code of the Neural Fast Downward planning system, along with the dataset used in Section~\ref{sec:experiment2}. This study was financed in part by the \textit{Coordenação de Aperfeiçoamento de Pessoal de Nível Superior – Brasil} (CAPES) – Finance Code 001. We acknowledge support from FAPERGS with project 21/2551-0000741-9. M.~R.~acknowledges support from CNPq (grants 437859/2018-5 and 310259/2022-3).

\printbibliography

\end{document}